\documentclass{article}
\pdfoutput=1

\usepackage{times}

\usepackage{graphicx} 

\usepackage{subfigure}

\usepackage{natbib}

\usepackage{algorithm}
\usepackage{algorithmic}

\usepackage{hyperref}

\usepackage[accepted]{style/icml2016}

\icmltitlerunning{Opponent Modeling in Deep Reinforcement Learning}

\newif\ifcomment\commentfalse

\usepackage{mdwlist}
\usepackage{latexsym}
\usepackage{nicefrac}
\usepackage{booktabs}
\usepackage{amsfonts}
\usepackage{bold-extra}
\usepackage{amsmath}
\usepackage{amssymb}
\usepackage{bm}
\usepackage{xcolor}
\usepackage{comment}
\usepackage{tabularx}
\usepackage{environ}
\usepackage{dsfont}
\usepackage{textcomp}
\usepackage{float}
\usepackage{graphicx}
\usepackage{mathtools}
\usepackage{microtype}
\usepackage{multirow}
\usepackage{multicol}
\usepackage{url}
\usepackage{latexsym}

\newcommand{\act}[1]{{\small \texttt{#1}}}

\newcommand{\gem}[1]{\mbox{\textsc{gem}}}
\newcommand{\abr}[1]{\textsc{#1}}

\newcommand{\e}[2]{\mathbb{E}_{#1}\left[ #2 \right] }

\newcommand{\hidetext}[1]{}
\newcommand{\ignore}[1]{}

\ifcomment
\newcommand{\nahocomment}[1]{
  \colorbox{yellow}{   \parbox{.8\linewidth}{\scriptsize NAHO: #1}  }}
\newcommand{\shcomment}[1]{ \colorbox{green}{  \parbox{.8\linewidth}{ SH:  #1}}}
\newcommand{\mjpcomment}[1]{   \colorbox{blue}{MJP: #1}}
\newcommand{\jbgcomment}[1]{  \colorbox{red}{   \parbox{.8\linewidth}{ JBG: #1}  }}
\newcommand{\fpcomment}[1]{  \colorbox{green}{   \parbox{.8\linewidth}{ FP: #1}  }}

\newcommand{\yhcomment}[1]{  \colorbox{green}{  \parbox{.8\linewidth}{ YH:  #1}
  }}
\newcommand{\hhecomment}[1]{  \colorbox{blue}{  \parbox{.8\linewidth}{ HH:  #1}
  }}
\newcommand{\jjmcomment}[1]{  \colorbox{green}{  \parbox{.8\linewidth}{ John:  #1}
  }}
\newcommand{\tncomment}[1]{  \colorbox{blue}{  \parbox{.8\linewidth}{ TN:  #1}
  }}
\newcommand{\mnicomment}[1]{  \colorbox{green}{  \parbox{.8\linewidth}{ Mohit:  #1}  }}

\newcommand{\jszcomment}[1]{  \colorbox{green}{  \parbox{.8\linewidth}{ JSG:  #1}  }}
\newcommand{\ascomment}[1]{  \colorbox{blue}{  \parbox{.8\linewidth}{ AS:  #1}
  }}
\newcommand{\vecomment}[1]{  \colorbox{blue}{  \parbox{.8\linewidth}{ VE:  #1}  }}
\newcommand{\halcomment}[1]{  \colorbox{magenta!20}{  \parbox{.8\linewidth}{ Hal:  #1}  }}
\newcommand{\kgcomment}[1]{  \colorbox{blue}{  \parbox{.8\linewidth}{ Kim:  #1}  }}
\newcommand{\vancomment}[1]{
  \colorbox{green}{  \parbox{.8\linewidth}{ VAN:  #1}  }}

\newcommand{\alvincomment}[1]{  \colorbox{cyan}{  \parbox{.8\linewidth}{ Alvin:  #1}  }}
\newcommand{\reviewercomment}[1]{  \colorbox{blue}{  \parbox{.8\linewidth}{Reviewer:  #1}  }}
\newcommand{\brscomment}[1]{  \colorbox{blue}{  \parbox{.8\linewidth}{BRS:  #1}  }}
\newcommand{\psrcomment}[1]{  \colorbox{yellow}{  \parbox{.8\linewidth}{PSR:  #1}  }}
\newcommand{\zkcomment}[1]{  \colorbox{cyan}{  \parbox{.8\linewidth}{ZK:  #1}  }}
\newcommand{\ctcomment}[1]{
  \colorbox{blue}{  \parbox{.8\linewidth}{CT:  #1}  }}
\newcommand{\swcomment}[1]{ \colorbox{yellow}{ \parbox{.8\linewidth}{ SW: #1}
  }}
\newcommand{\shaycomment}[1]{  \colorbox{yellow}{  \parbox{.8\linewidth}{SBC:  #1}  }}
\newcommand{\jlundcomment}[1]{  \colorbox{cyan}{  \parbox{.8\linewidth}{J:  #1}  }}
\newcommand{\kdscomment}[1]{  \colorbox{ceil}{  \parbox{.8\linewidth}{KDS:  #1}  }}
\newcommand{\lkfcomment}[1]{  \colorbox{yellow}{  \parbox{.8\linewidth}{LF:  #1}  }}
\else
\newcommand{\nahocomment}[1]{ }
\newcommand{\shcomment}[1]{ }
\newcommand{\alvincomment}[1]{ }
\newcommand{\jbgcomment}[1]{ }
\newcommand{\yhcomment}[1]{ }
\newcommand{\jjmcomment}[1]{ }
\newcommand{\hhecomment}[1]{ }
\newcommand{\tncomment}[1]{ }
\newcommand{\mnicomment}[1]{ }
\newcommand{\ascomment}[1]{ }
\newcommand{\vecomment}[1]{ }
\newcommand{\halcomment}[1]{ }
\newcommand{\kgcomment}[1]{ }
\newcommand{\brscomment}[1]{ }
\newcommand{\reviewercomment}[1]{ }
\newcommand{\zkcomment}[1]{ }
\newcommand{\jszcomment}[1]{ }
\newcommand{\ctcomment}[1]{ }
\newcommand{\swcomment}[1]{ }
\newcommand{\psrcomment}[1]{ }
\newcommand{\vancomment}[1]{ }
\newcommand{\shaycomment}[1]{ }
\newcommand{\jlundcomment}[1]{ }
\newcommand{\kdscomment}[1]{ }
\newcommand{\lkfcomment}[1]{ }
\newcommand{\mjpcomment}[1]{ }
\newcommand{\fpcomment}[1]{ }
\fi

\newcommand{\smallurl}[1]{ \begin{scriptsize}\url{#1}\end{scriptsize}}

\NewEnviron{smalign}{
\vspace{-.6cm}
\begin{small}
\begin{align}
  \BODY
\end{align}
\end{small}
\vspace{-.6cm}
}

\definecolor{CUgold}{HTML}{CFB87C}
\definecolor{grey}{rgb}{0.95,0.95,0.95}
\definecolor{ceil}{rgb}{0.57, 0.63, 0.81}

\usepackage{breqn}
\newcommand{\dqn}{\abr{dqn}}
\newcommand{\dron}{\abr{dron}}
\newcommand{\dronmoe}{\abr{dron-moe}}
\newcommand{\gru}{\abr{gru}}
\newcommand{\oa}{o}
\newcommand{\op}{\pi^o}

\begin{document}

\twocolumn[
\icmltitle{Opponent Modeling in Deep Reinforcement Learning}

\icmlauthor{He He}{hhe@umiacs.umd.edu}
\icmladdress{University of Maryland,
            College Park, MD 20740 USA}
\icmlauthor{Jordan Boyd-Graber}{Jordan.Boyd.Graber@colorado.edu}
\icmladdress{University of Colorado,
            Boulder, CO 80309 USA}
\icmlauthor{Kevin Kwok}{kkwok@mit.edu}
\icmladdress{Massachusetts Institute of Technology,
            Cambridge, MA 02139 USA}
\icmlauthor{Hal Daum\'e III}{hal@umiacs.umd.edu}
\icmladdress{University of Maryland,
            College Park, MD 20740 USA}

\icmlkeywords{opponent modeling, reinforcement learning, neural network}

\vskip 0.3in
]

\begin{abstract}

  Opponent modeling is necessary in multi-agent settings where
  secondary agents with competing goals also adapt their strategies,
  yet it remains challenging because strategies interact with each
  other and change.  Most previous work focuses on
  developing probabilistic models or parameterized strategies for
  specific applications.  Inspired by the recent success of deep
  reinforcement learning, we present neural-based models that jointly
  learn a policy and the behavior of opponents.  Instead of explicitly
  predicting the opponent's action, we encode observation of the
  opponents into a deep Q-Network (\abr{dqn}); however, we retain
  explicit modeling (if desired) using multitasking.  By using a Mixture-of-Experts
  architecture, our model automatically discovers different strategy
  patterns of opponents without extra supervision.  We
  evaluate our models on a simulated soccer game and a popular trivia
  game, showing superior performance over \abr{dqn} and its variants.
\end{abstract}

\section{Introduction}
\label{sec:intro}

An intelligent agent working in strategic settings (e.g., collaborative or
competitive tasks) must predict the action of other agents and reason about
their intentions.  This is important because all active agents affect the state
of the world.  For example, a multi-player game \abr{ai} can exploit suboptimal
players if it can predict their bad moves; a negotiating agent can reach an
agreement faster if it knows the other party's bottom line; a self-driving car
must avoid accidents by predicting where cars and pedestrians are going.  Two
critical questions in opponent modeling are what variable(s) to model and how to
use the predicted information.  However, the answers depend much on the specific
application, and most previous
work~\cite{opponent-modeling-in-poker,bayesbluff,game-theory-opponent-modeling}
focuses exclusively on poker games which require substantial domain knowledge.

We aim to build a general opponent modeling framework in the
reinforcement learning setting, which enables the agent to exploit
idiosyncrasies of various opponents.  First, to account for
changing behavior, we model uncertainty in the opponent's
strategy instead of classifying it into a set of stereotypes.
Second, domain knowledge is often required when prediction of the
opponents are separated from learning the dynamics of the world.
Therefore, we jointly learn a policy and model the opponent
probabilistically.

We develop a new model, \dron{} (Deep
Reinforcement Opponent Network), based on the recent deep
Q-Network of \citet[\dqn{}]{mnih-dqn-2015} in Section~\ref{sec:method}.
\dron{} has a policy learning module that predicts Q-values
and an opponent learning module that infers opponent strategy.\footnote{Code and data: \url{https://github.com/hhexiy/opponent}}
Instead of explicitly predicting
opponent properties, \dron{} learns hidden representation of the opponents
based on past observations and uses it (in addition to the state information) to compute an adaptive response.
More specifically,
we propose two architectures, one using simple concatenation to combine the two modules and one based on the Mixture-of-Experts network.
While we model opponents implicitly,
additional supervision (e.g., the action or strategy taken)
can be added through multitasking.

Compared to previous models
that are specialized in particular applications, \dron{} is designed with a
general purpose and does not require knowledge of possible (parameterized)
game strategies.

A second contribution is \dqn{} agents that learn in multi-agent settings. Deep
reinforcement learning has shown competitive performance in various tasks:
arcade games~\cite{mnih-dqn-2015}, object
recognition~\cite{mnih14visualattention}, and robot
navigation~\cite{zhang15dnnmemory}.  However, it has been mostly applied to the
single-agent decision-theoretic settings with stationary environments.  One
exception is \citet{multiagent-dqn}, where two agents controlled by independent
\dqn{}s interact under collaborative and competitive rewards.  While their focus
is the collective behavior of a multi-agent system with known controllers, we
study from the view point of a single agent that must learn a reactive
policy in a stochastic environment filled with \emph{unknown}
opponents.

We evaluate our method on two tasks in Section~\ref{sec:experiments}: a
simulated two-player soccer game in a grid world, and a real question-answering
game, quiz bowl, against users playing online.  Both games have opponents with a mixture of
strategies that require different counter-strategies.  Our model consistently
achieves better results than the \dqn{} baseline.  In addition, we show our
method is more robust to non-stationary strategies; it successfully identifies
the opponent's strategy and responds correspondingly.

\section{Deep Q-Learning}
\label{sec:background}

Reinforcement learning is commonly used for solving Markov-decision
processes (\abr{mdp}), where an agent interacts with the world and
collects rewards.  Formally, the agent takes an action $a$ in state
$s$, goes to the next state $s'$ according to the transition
probability $\mathcal{T}(s,a,s')=Pr(s'|s,a)$ and receives reward $r$.
States and actions are defined by the state space $\mathcal{S}$ and
the action space $\mathcal{A}$.  Rewards $r$ are assigned by a real-valued
reward function $\mathcal{R}(s, a, s')$.  The agent's behavior is
defined by a policy $\pi$ such that $\pi(a|s)$ is the probability of
taking action $a$ in state $s$.  The goal of reinforcement learning is
to find an optimal policy $\pi^\ast$ that maximizes the expected
discounted cumulative
reward 
$R = \e{}{\sum_{t=0}^T\gamma^t r_t}$, where $\gamma \in [0, 1]$ is the
discount factor and $T$ is the time step when the episode ends.

One approach to solve \abr{mdp}s is to compute its $Q$-function:
the expected reward starting from state $s$, taking action $a$ and following policy $\pi$:

$Q^\pi(s,a) \equiv \e{}{\sum_{t} \gamma^t r_t | s_0=s, a_0=a, \pi}$.  Q-values
of an optimal policy solve the Bellman Equation~\cite{rl-intro}:
\begin{equation*}
\label{eqn:qiteration}
Q^\ast(s,a) = \sum_{s'}\mathcal{T}(s,a,s')\left [ r + \gamma \max_{a'} Q^\ast(s', a')\right ].
\end{equation*}
Optimal policies always select the action with the highest Q-value for a given
state.  Q-learning~\cite{qlearning,rl-intro}
finds the optimal Q-values without knowledge of $\mathcal{T}$.
Given observed transitions $(s, a, s', r)$, Q-values are updated recursively:
\begin{equation*}
\label{eqn:qlearning}
Q(s,a) \leftarrow Q(s,a) + \alpha\left [r + \gamma\max_{a'} Q(s', a') - Q(s,a)\right ].
\end{equation*}

For complex problems with continuous states, the $Q$-function cannot be
expressed as a lookup table, requiring a continuous approximation.  Deep
reinforcement learning such as \dqn{}~\cite{mnih-dqn-2015}---a deep Q-learning
method with experience replay---approximates the $Q$-function using a neural
network.  It draws samples $(s, a, s', r)$ from a replay memory $M$, and the
neural network predicts $Q^\ast$ by minimizing squared loss at iteration
$i$:
\begin{dmath*}
L^i(\theta^i) = \mathbb{E}_{(s, a, s', r) \sim U(M)} \left [ \left ( {r + \gamma\max_{a'}Q(s',a';\theta^{i-1})} - Q(s,a;\theta^i)\right ) ^2 \right ],
\end{dmath*}
where $U(M)$ is a uniform distribution over replay memory.

\begin{figure}[t]
\centering
\includegraphics[width=0.42\textwidth]{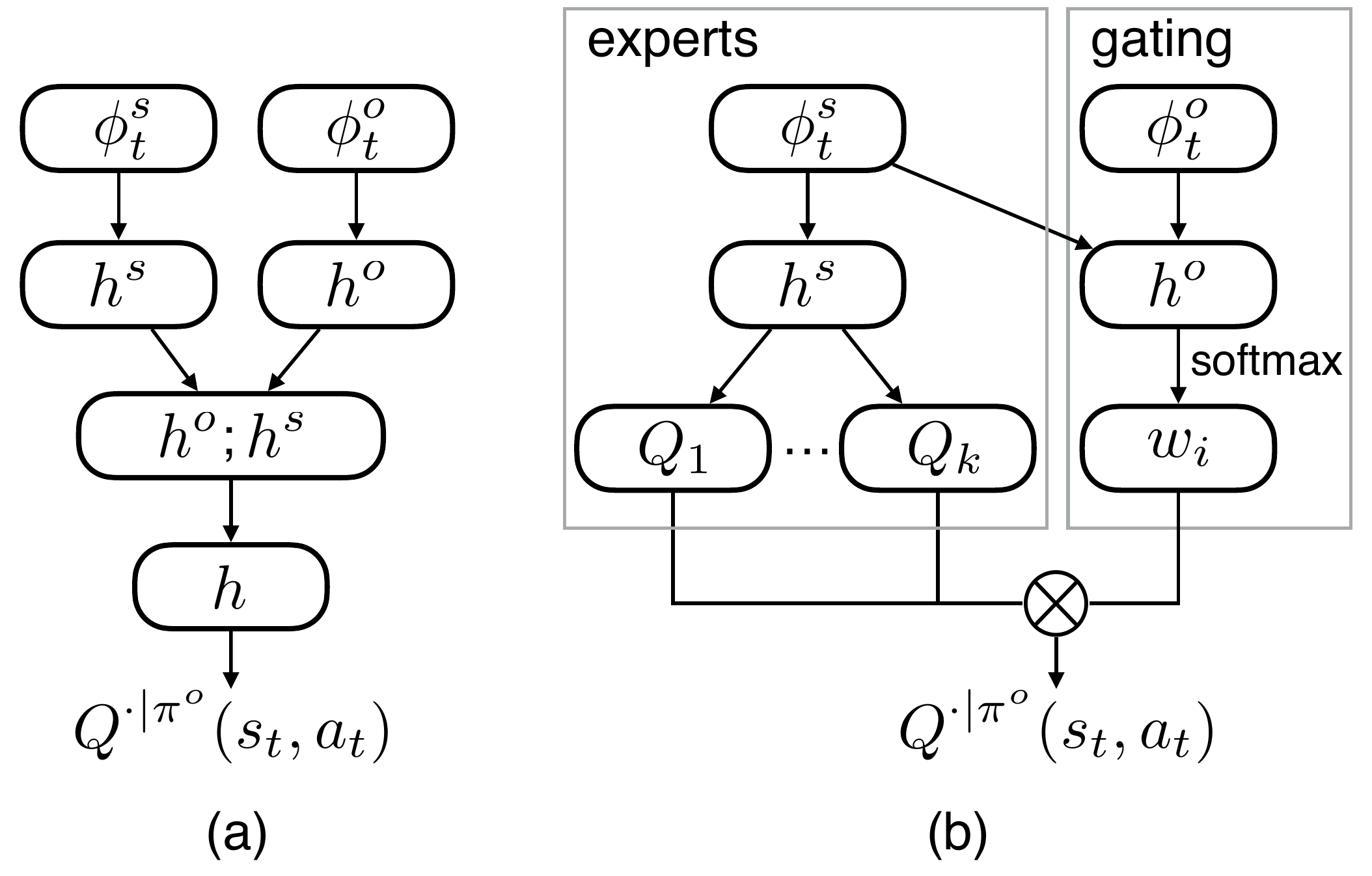}
\caption{Diagram of the \dron{} architecture. (a) \dron{}-concat: opponent representation is concatenated with the state representation.
(b) \dron{}-MoE: Q-values predicted by $K$ experts are combined linearly by weights from the gating network.}
\label{fig:dron}
\end{figure}

\begin{figure}[t]
\centering
\includegraphics[width=0.48\textwidth]{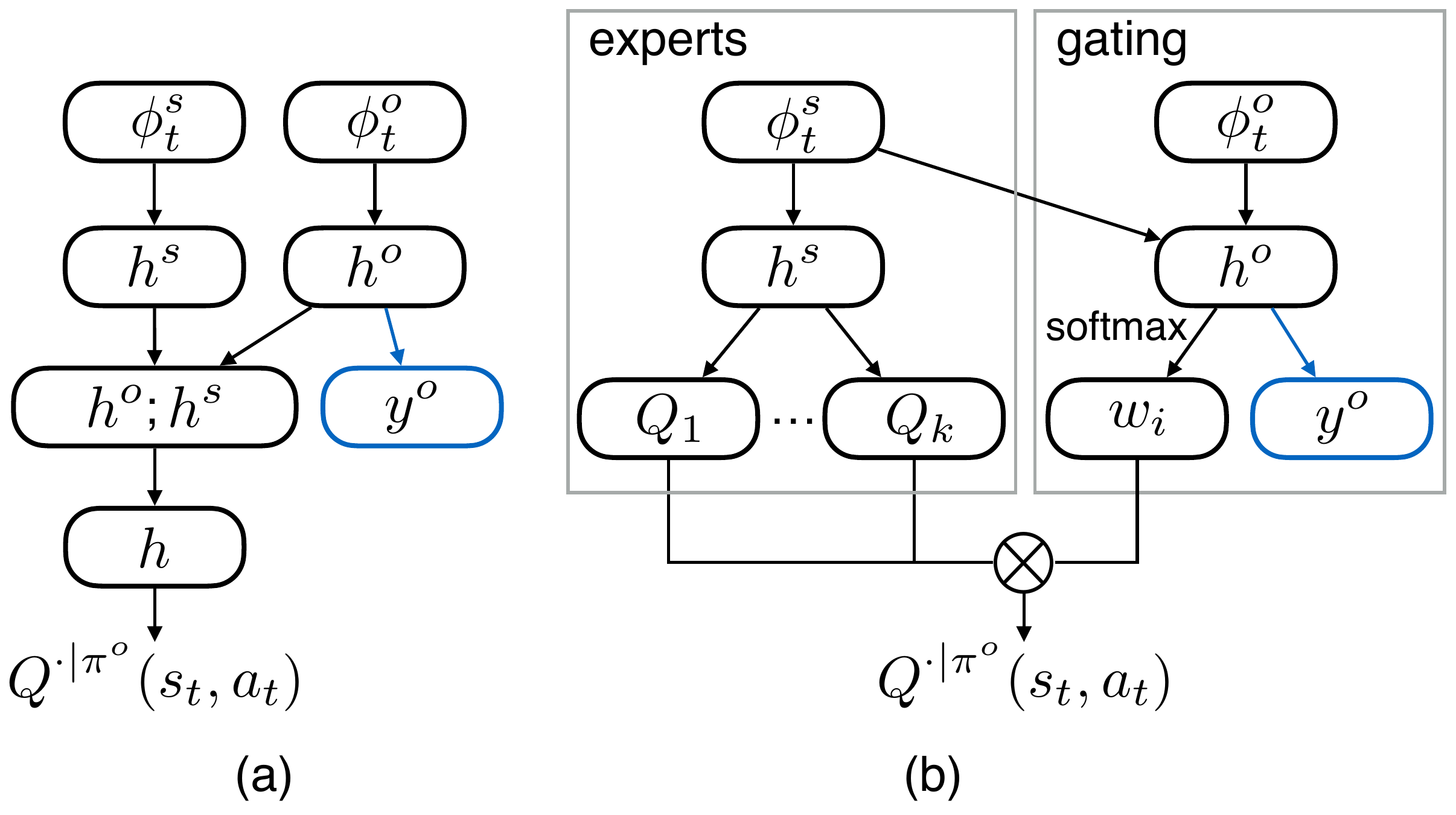}
\caption{Diagram of the DRON with multitasking. The blue part shows that the supervision signal from the opponent affects the Q-learning network by changing the opponent features.}
\label{fig:dron-mt}
\end{figure}

\section{Deep Reinforcement Opponent Network}
\label{sec:method}

In a multi-agent setting, the environment is affected by the \emph{joint} action
of all agents.  From the perspective of one agent, the outcome of an action in a
given state is no longer stable, but is dependent on actions of other agents.
In this section, we first analyze the effect of multiple agents on the Q-learning framework;
then we present \dron{} and its multitasking variation.

\subsection{Q-Learning with Opponents}

In \abr{mdp} terms, the joint action space is defined by $\mathcal{A}^M =
\mathcal{A}_1 \times \mathcal{A}_2 \times \ldots \times \mathcal{A}_n$ where $n$
is the total number of agents.  We use $a$ to denote the action of the agent we
control (the primary agent) and $\oa$ to denote the joint action of all other agents (secondary agents), such that $(a,
\oa) \in \mathcal{A}^M$.  Similarly, the transition probability becomes
$\mathcal{T}^M(s,a,\oa,s')=Pr(s'|s,a,\oa)$, and the new reward function is
$\mathcal{R}^M(s,a,\oa,s')$.  Our goal is to learn an optimal policy for the primary
agent given interactions with the joint policy $\op$ of the secondary
agents.\footnote{While a joint policy defines the distribution of joint actions,
  the opponents may be controlled by independent policies.}

If $\op$ is stationary, then the multi-agent \abr{mdp} reduces to a single-agent
\abr{mdp}: the opponents can be considered part of the world.  Thus, they
redefine the transitions and reward:
\begin{eqnarray*}
\mathcal{T}(s,a,s')&=&\sum_{\oa} \op(\oa|s) \mathcal{T}^M(s,a,\oa,s'), \\
\mathcal{R}(s,a,s')&=&\sum_{\oa} \op(\oa|s) \mathcal{R}^M(s,a,\oa,s').
\end{eqnarray*}
Therefore, an agent can ignore other agents,
and standard Q-learning suffices.

Nevertheless, it is often unrealistic to assume opponents use fixed
policies.  Other agents may also be learning or adapting to maximize rewards.
For example, in strategy games, players may disguise their true
strategies at the beginning to fool the opponents; winning players
protect their lead by playing defensively; and losing players play
more aggressively.  In these situations, we face opponents with an unknown
policy $\op_t$ that changes over time.

Considering the effects of other agents, the definition of an optimal policy in
Section~\ref{sec:background} no longer applies---the effectiveness policies now
depends on policies of secondary agents.  We therefore define the optimal
$Q$-function relative to the joint policy of opponents: $Q^{\ast|\op} =
\max_{\pi} Q^{\pi|\op}(s,a) \; \forall s \in \mathcal{S} \;\mbox{and}\; \forall
a \in \mathcal{A}$.  The recurrent relation between Q-values holds:
\begin{eqnarray}
Q^{\pi|\op}(s_t,a_t) = \sum_{\oa_t} \op_t(\oa_t|s_t) \sum_{s_{t+1}} \mathcal{T}(s_t,a_t,\oa_t,s_{t+1}) \nonumber \\
\left[ \mathcal{R}(s_t,a_t,\oa_t,s_{t+1}) + \gamma\e{a_{t+1}}{Q^{\pi|\op}(s_{t+1},a_{t+1})} \right] .
\label{eqn:opponent-q}
\end{eqnarray}

\subsection{DQN with Opponent Modeling}

Given Equation~\ref{eqn:opponent-q}, we can continue applying Q-learning and
estimate both the transition function and the opponents' policy by stochastic
updates.  However, treating opponents as part of the world can slow
responses to adaptive opponents~\cite{opponent-qlearning}, because
the change in behavior is masked by the dynamics of the world.

To encode opponent behavior explicitly, we propose the Deep
Reinforcement Opponent Network (\dron{}) that models $Q^{\cdot|\op}$
and $\op$ jointly.  \dron{} is a Q-Network ($N_Q$) that
evaluates actions for a state and an opponent network ($N_o$) that
learns representation of $\op$.  The remaining questions are how to
combine the two networks and what supervision signal to use.  To
answer the first question, we investigate two network architectures:
\dron{}-concat that concatenates $N_Q$ and $N_o$, and \dronmoe{} that
applies a Mixture-of-Experts model.

To answer the second question, we consider two settings:
(a) predicting Q-values only, as our goal is the best reward instead of accurately simulating opponents;
and (b) also predicting extra information about the opponent when it is available, e.g., the type of their strategy.

\paragraph{\dron-concat}

We extract features from the state ($\phi^s$) and the opponent ($\phi^o$ ) and
then use linear layers with rectification or convolutional neural networks---$N_Q$ and $N_o$---to embed them
in separate hidden spaces ($h^s$ and $h^o$).
To incorporate knowledge of $\op$ into the Q-Network,
we concatenate representations of the
state and the opponent (Figure~\ref{fig:dron}a).  The
concatenation then jointly predicts the Q-value.  Therefore, the last layer(s) of the neural
network is responsible for understanding the interaction between opponents and
Q-values.  Since there is only one Q-Network, the model requires a more discriminative representation of the opponents to learn an adaptive policy.
To alleviate this, our second model encodes a
stronger prior of the relation between opponents' actions and Q-values based on Equation~\ref{eqn:opponent-q}.

\paragraph{\dronmoe}

The right part of Equation~\ref{eqn:opponent-q} can be written as
$\sum_{\oa_t} \op_t(\oa_t|s_t) Q^\pi(s_t,a_t,\oa_t)$, an expectation
over different opponent behavior.  We use a
Mixture-of-Experts network~\cite{moe} to explicitly model the opponent
action as a hidden variable and marginalize over it
(Figure~\ref{fig:dron}b).  The expected Q-value is obtained by
combining predictions from multiple \emph{expert networks}:
\begin{eqnarray*}
Q(s_t, a_t; \theta) &=& \sum_{i=1}^K w_i Q_i(h^s, a_t) \\
Q_i(h^s, \cdot) &=& f(W_i^s h^s + b_i^s) .
\end{eqnarray*}
Each expert network predicts a possible reward in the current state.
A \emph{gating network} based on the opponent representation computes combination weights (distribution over experts):
\begin{equation*}
w = \mbox{softmax}\left(f(W^o h^o + b^o)\right).
\end{equation*}
Here $f(\cdot)$ is a nonlinear activation function (ReLU for
all experiments),
$W$ represents the linear transformation matrix, and $b$ is the bias term.

Unlike \dron{}-concat, which ignores the interaction between the world and
opponent behavior, \dronmoe{} knows that Q-values have different distributions
depending on $\phi^o$; each expert network captures one type of
opponent strategy.

\paragraph{Multitasking with \dron{}}

The previous two models
predict Q-values only, thus the opponent representation is learned
indirectly through feedback from the Q-value.  Extra information
about the opponent can provide direct supervision for $N_o$.
Many games reveal additional information besides the
final reward at the end of a game.  At the very least the agent has
observed actions taken by the opponents in past states; sometimes
their private information such as the hidden cards in poker.  More high-level
information includes abstracted plans or strategies.
Such information reflects characteristics of opponents and
can aid policy learning.

Unlike previous work that learns a separate model to predict these
information about the
opponent~\cite{davidson99opponent,game-theory-opponent-modeling,schadd07opponentmodeling},
we apply multitask learning and use the observation as extra
supervision to learn a \emph{shared} opponent representation
$h^o$.  Figure~\ref{fig:dron-mt} shows the architecture of multitask
\dron{}, where supervision is $y^o$.
The advantage of multitasking over explicit opponent modeling is that
it uses high-level knowledge of the game and the opponent, while
remaining robust to insufficient opponent data and modeling error
from Q-values.  In Section~\ref{sec:experiments},
we evaluate multitasking \dron{} with two types of supervision
signals: future action and overall strategy of the opponent.

\begin{figure}[t!]
  \begin{minipage}[b]{.23\textwidth}
    \includegraphics[width=1\linewidth]{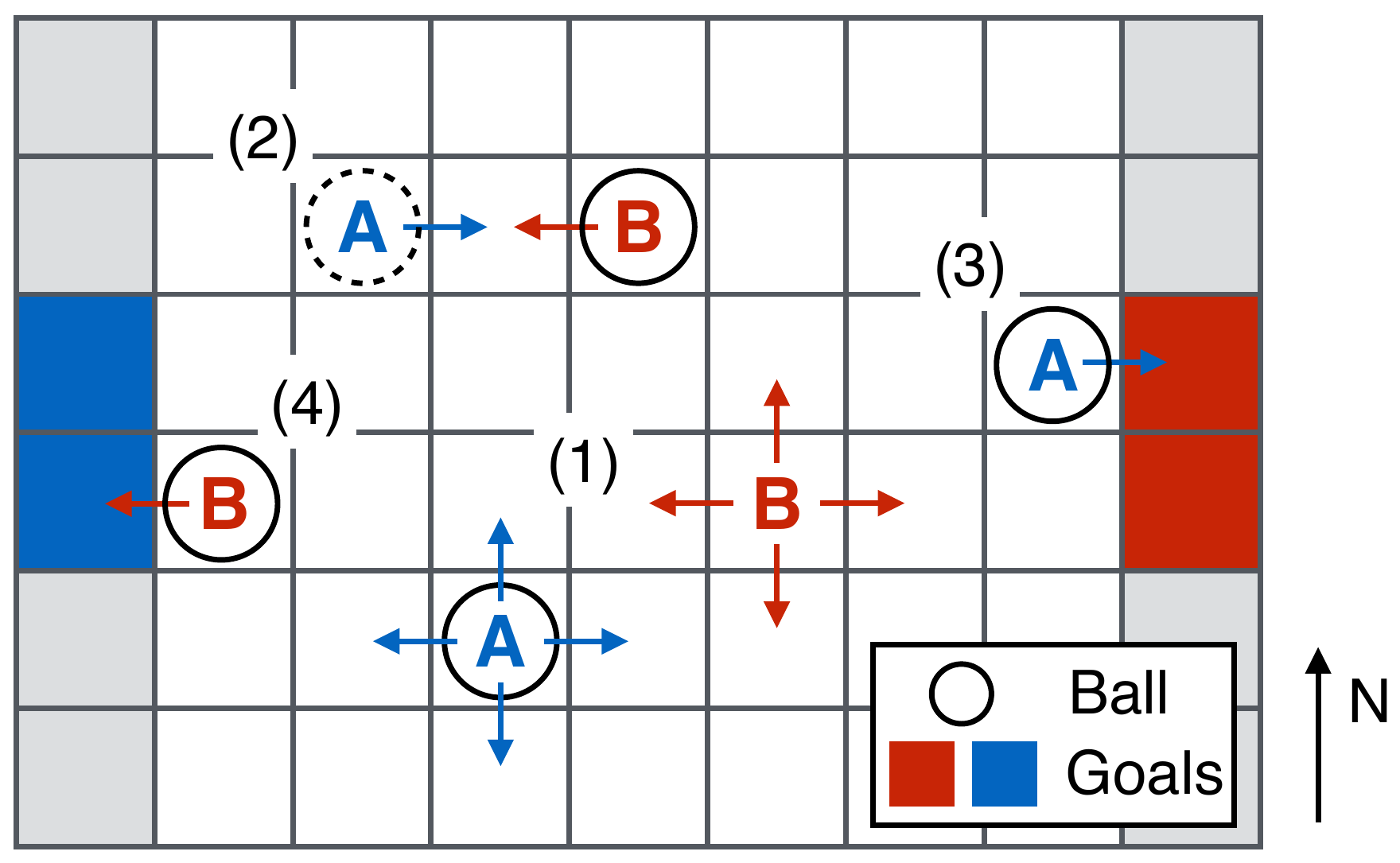}
  \end{minipage}
  \begin{minipage}[b]{.23\textwidth}
    \setlength{\tabcolsep}{1.5pt}
    \renewcommand{\arraystretch}{0.7}
    \begin{tabular}[b]{ccc}
      \toprule
      & Defensive & Offensive \\
      \midrule
      \multirow{2}{*}{w/ ball} & Avoid & Advance \\
      & opponent & to goal \vspace{2pt} \\
      \midrule\vspace{2pt}
      \multirow{2}{*}{w/o ball} & Defend & Intercept \\
      & goal & the ball \\
      \bottomrule
    \end{tabular}
  \end{minipage}
  \caption{\emph{Left:} Illustration of the soccer game.
    \emph{Right:} Strategies of the hand-crafted rule-based agent.}
  \label{fig:soccer_game}
\end{figure}

\section{Experiments}
\label{sec:experiments}

In this section, we evaluate our models on two tasks, the soccer game
and quiz bowl.  Both tasks have two players against each other and the
opponent presents varying behavior.  We compare \dron{} models with
\dqn{} and analyze their response against different types of
opponents.

All systems are trained under the same Q-learning framework.  Unless
stated otherwise, the experiments have the following
configuration: discount factor $\gamma$ is 0.9, parameters are optimized by AdaGrad~\cite{duchi2011adaptive} with a learning rate of 0.0005, and the mini-batch size is 64.  We use
$\epsilon$-greedy exploration during training, starting with an
exploration rate of 0.3 that linearly decays to 0.1 within 500,000
steps.  We train all models for fifty epochs.
Cross Entropy is used as the loss in multitasking learning.

\subsection{Soccer}

Our first testbed is a soccer variant following previous work on
multi-player
games~\cite{minimax-q,collins-thesis-rl,opponent-qlearning}.  The game
is played on a $6\times 9$ grid (Figure~\ref{fig:soccer_game}) by two
players, A and B.\footnote{Although the game is played in a grid
  world, we do not represent the $Q$-function in tabular form as in
  previous work. Therefore it can be generalized to more complex
  pixel-based settings.}  The game starts with A and B in a
randomly squares in the left and right half (except the
goals), and the ball goes to one of them.  Players choose
from five actions: \act{move N, S, W, E} or \act{stand} still
(Figure~\ref{fig:soccer_game}(1)).  An action is invalid if it takes
the player to a shaded square or outside of the border.  If two
players move to the same square, the player who possesses the ball
before the move loses it to the opponent
(Figure~\ref{fig:soccer_game}(2)), and the move does not take place.

A player scores one point if they take the ball
to the opponent's goal (Figure~\ref{fig:soccer_game}(3), (4)) and the
game ends.  If neither player gets a goal within one hundred steps,
the game ends with a zero--zero tie.

\paragraph{Implementation}

We design a two-mode rule-based agent as the opponent Figure~\ref{fig:soccer_game}(right).
In the offensive mode, the agent always prioritize
attacking over defending.  In 5000 games against a random agent, it wins 99.86\%
of the time and the average episode length is 10.46.
In defensive mode, the agent only
focuses on defending its own goal.  As a result, it wins 31.80\% of the games
and ties 58.40\% of them; the average episode length is 81.70.
It is easy to find a strategy to defeat the opponent in either mode,
however, the strategy does not work well for both modes, as we will show in Table~\ref{tab:soccer_matrix}.
Therefore, the agent randomly chooses between the two modes in each game to create a varying strategy.

The input state is a $1\times 15$ vector representing coordinates of
the agent, the opponent, the axis limits of the field, positions of
the goal areas and ball possession.  We define a player's move
by five cases: approaching the agent, avoiding the agent, approaching the agent's goal,
approaching self goal and standing still.  Opponent features
include frequencies of observed opponent moves, its most recent move
and action, and the frequency of losing the ball to the opponent.

The baseline \dqn{} has two hidden layers, both with 50 hidden units.
We call this model \dqn{}-world: opponents are modeled as
part of the world.  The hidden layer of the opponent network in
\dron{} also has 50 hidden units.  For multitasking, we experiment
with two supervision signals, opponent action in the current state
(+action) and the opponent mode (+type).

\paragraph{Results}
\begin{table}

\centering
\begin{tabular}[ht]{lccc}
\toprule
\multirow{2}{*}{Model} & \multirow{2}{*}{Basic} & \multicolumn{2}{c}{Multitask} \\
& & +action & +type \\
\midrule
\multicolumn{4}{c}{Max $R$} \\
\midrule
\dron{}-concat & 0.682 & 0.695$^\ast$ & 0.690$^\ast$ \\
\dronmoe{} & {\bf 0.699}$^\ast$ & 0.697$^\ast$ & 0.686$^\ast$ \\
\dqn{}-world & 0.664 & - & - \\
\midrule
\multicolumn{4}{c}{Mean $R$} \\
\midrule
\dron{}-concat & 0.660 & 0.672 & 0.669 \\
\dronmoe & 0.675  & 0.664 & 0.672 \\
\dqn{}-world & 0.616 & - & - \\
\bottomrule
\end{tabular}
\caption{
  Rewards of \dqn{} and \dron{} models on Soccer.
  We report the maximum test reward ever achieved (Max $R$) and
  the average reward of the last 10 epochs (Mean $R$).
  Statistically significant ($p<0.05$ in two-tailed pairwise
  $t$-tests) improvement for \dqn{} ($^\ast$) and all other
  models in {\bf bold}. \dron{} models achieve higher rewards in both measures.
}

\label{tab:soccer_model_result}
\end{table}

\begin{figure}
\centering
\includegraphics[width=0.6\linewidth]{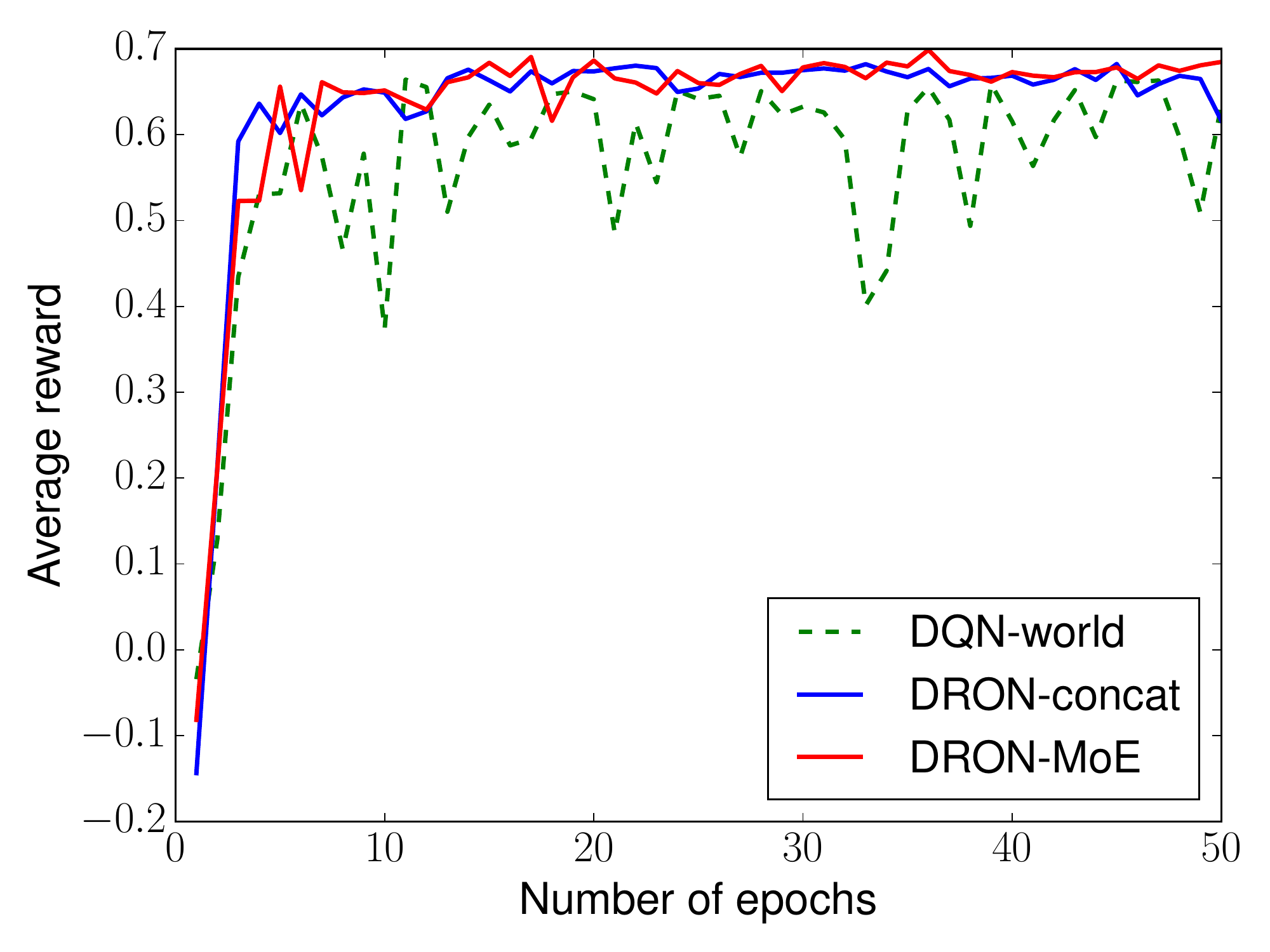}
\caption{Learning curves on Soccer over fifty epochs. \dron{} models are
  more stable than \dqn{}.}
\label{fig:soccer_all_test_rewards}
\end{figure}

In Table~\ref{tab:soccer_model_result}, we compare rewards of \dron{}
models, their multitasking variations, and \dqn{}-world.  After each
epoch, we evaluate the policy with 5000 randomly generated games (the test set) and
compute the average reward.  We report the mean test reward after the
model stabilizes and the maximum test reward ever achieved.  The
\dron{} models outperform the \dqn{} baseline.  Our model also has much
smaller variance (Figure~\ref{fig:soccer_all_test_rewards}).

Adding additional
supervision signals improves \dron-concat but not \dronmoe\ (multitask column).
\dron-concat does not explicitly learn
different strategies for different types of opponents, therefore more
discriminative opponent representation helps model the relation
between opponent behavior and Q-values.  However, for \dronmoe,
while better opponent representation is still desirable, the
supervision signal may not be aligned with ``classification'' of the
opponents learned from the Q-values.

\begin{table}[t!]
\centering
\begin{tabular}{cccccc}
\toprule
& \multicolumn{2}{c}{\dqn{}} & \dqn{}& \dron{} & \dron{} \\
& O only & D only & -world & -concat & -\abr{moe} \\
\midrule
O & {\bf 0.897} & -0.272 & 0.811 & 0.875 & 0.870 \\
D & 0.480 & {\bf 0.504} & 0.498 & 0.493 & 0.486 \\
\bottomrule
\end{tabular}
\caption{Average rewards of \dqn{} and \dron{} models when playing against different types of opponents. Offensive and defensive agents are represented by O and D. ``O only'' and ``D only'' means training against O and D agents only. Upper bounds of rewards are in bold. \dron{} achieves rewards close to the upper bounds against both types of opponents.}
\label{tab:soccer_matrix}
\end{table}

To investigate how the learned policies adapt to different opponents,
we test the agents against a defensive opponent and an offensive
opponent separately.  Furthermore, we train two \dqn{} agents targeting
at each type of opponent respectively.
Their performance is best an agent can do when facing a single type of opponent (in our setting),
as the strategies are learned to defeat this particular opponent.
Table~\ref{tab:soccer_matrix} shows the
average rewards of each model and the \dqn{} upper bounds (in bold).  \dqn{}-world is confused
by the defensive behavior and significantly sacrifices its performance
against the offensive opponent; \dron{} achieves a much better
trade-off, retaining rewards close to both upper bounds against the
varying opponent.

\begin{figure}[t]
\centering
\includegraphics[width=0.6\linewidth]{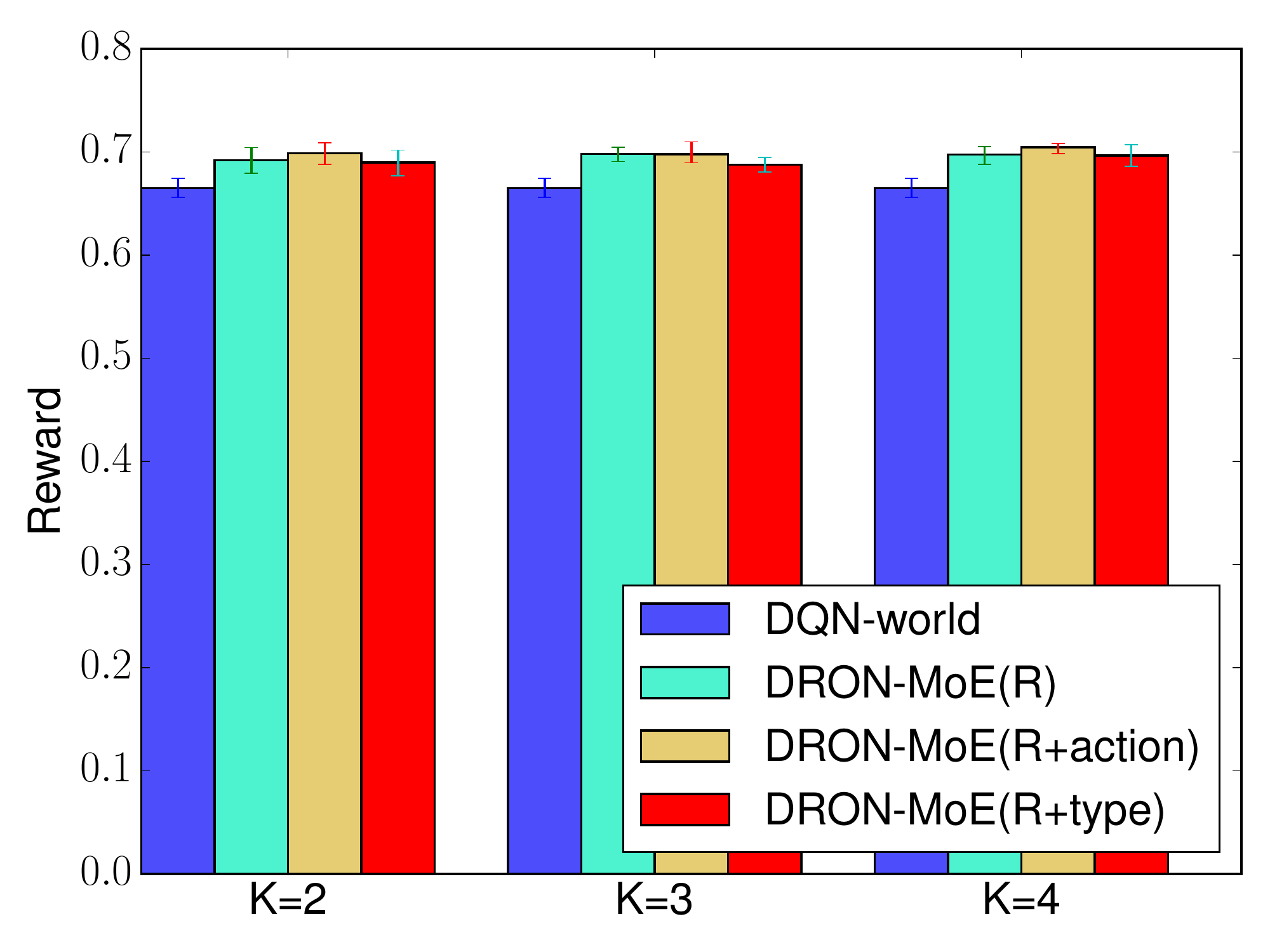}
\caption{Effect of varying the number experts (2--4) and multitasking on Soccer.
The error bars show the 90\% confidence interval.
\dronmoe{} consistently improves over DQN regardless of the number of mixture components.
Adding extra supervision does not obviously improve the results.
}
\label{fig:soccer_moe_bar}
\end{figure}

Finally, we examine how the number of experts in \dronmoe{} affects
the result. From Figure~\ref{fig:soccer_moe_bar}, we see no
significant difference in varying the number of experts, and
\dronmoe{} consistently performs better than \dqn{} across all $K$.
Multitasking does not help here.

\subsection{Quiz Bowl}

\begin{figure}[t]
\centering
\includegraphics[width=1.0\columnwidth]{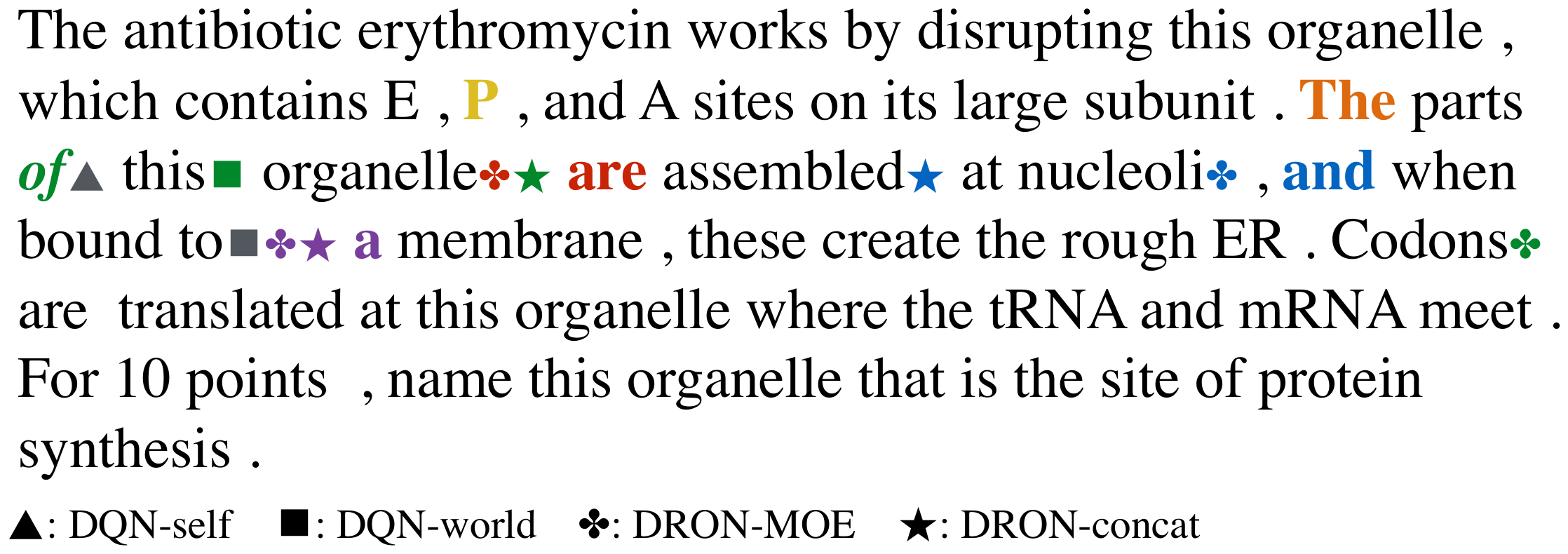}
\caption{Buzz positions of human players and agents on one science question whose answer is ``ribosome''.
Words where a player buzzes is displayed in a color unique to the player;
a wrong buzz is shown in \emph{italic}.
Words where an agent buzzes is subscripted by a symbol unique to the agent;
color of the symbol corresponds to the player it is playing against.
A gray symbol means that buzz position of the agent does not depend on its opponent.
\dron{} agents adjust their buzz positions according to the opponent's buzz position and correctness.
Best viewed in color.
}
\label{fig:buzz_example}
\end{figure}

Quiz bowl is a trivia game widely played in English-speaking countries between
schools, with tournaments held most
weekends. It is usually played between two
teams.  The questions are read to players and they score points by ``buzzing in''
first (often before the question is finished) and answering the question
correctly.
One example question with buzzes is shown in Figure~\ref{fig:buzz_example}.
A successful quiz bowl player needs two things: a content
model to predict answers given (partial) questions and a buzzing model to decide
when to buzz.

\paragraph{Content Model}

We model the question answering part as an incremental text-classification
problem.  Our content model is a recurrent neural network with gated recurrent
units (\gru{}).  It reads in the question sequentially and outputs a distribution
over answers at each word given past information encoded in the hidden states.

\paragraph{Buzzing Model}

To test depth of knowledge, questions start
with obscure information and reveals more and more obvious clues towards the
end (e.g., Figure~\ref{fig:buzz_example}).  Therefore, the buzzing model faces a speed-accuracy tradeoff: while
buzzing later increases one's chance of answering correctly, it also increases
the risk of losing the chance to answer.  A safe strategy is to always buzz as
soon as the content model is confident enough.  A smarter strategy, however, is
to adapt to different opponents: if the opponent often buzzes late, wait for more clues; otherwise, buzz more aggressively.

To model interaction with other players, we take a reinforcement learning
approach to learn a buzzing policy.  The state includes words revealed and
predictions from the content model, and the actions are \act{buzz} and
\act{wait}.  Upon buzzing, the content model outputs the most likely answer at
the current position.  An episode terminates when one player buzzes and answers
the question correctly.  Correct answers are worth 10 points and wrong answers
are $-5$.

\begin{figure}[t]
\centering
\subfigure[]{
\includegraphics[width=0.23\textwidth]{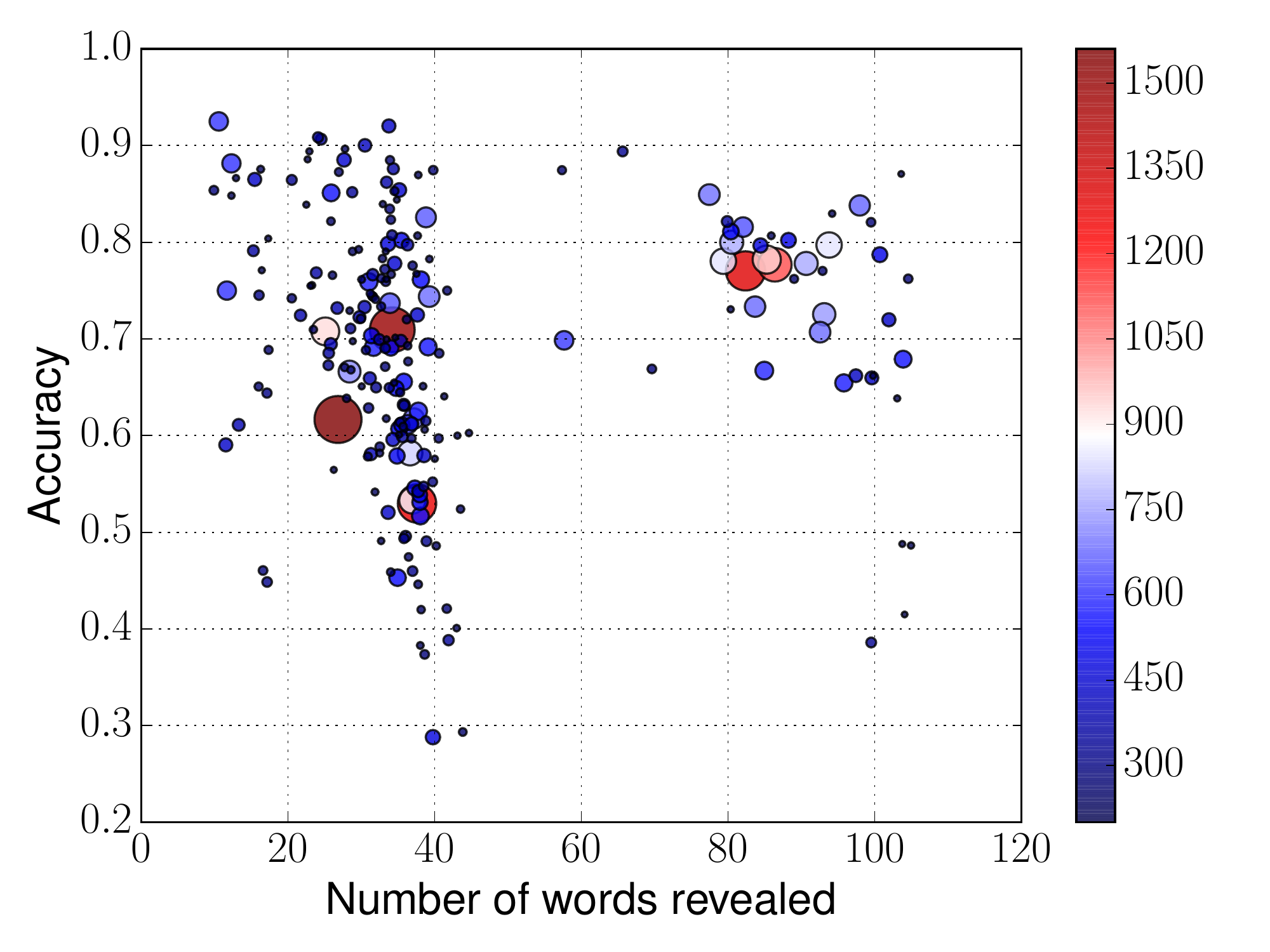}
\label{fig:player_acc}
}
\hspace{-1em}
\subfigure[]{
\includegraphics[width=0.23\textwidth]{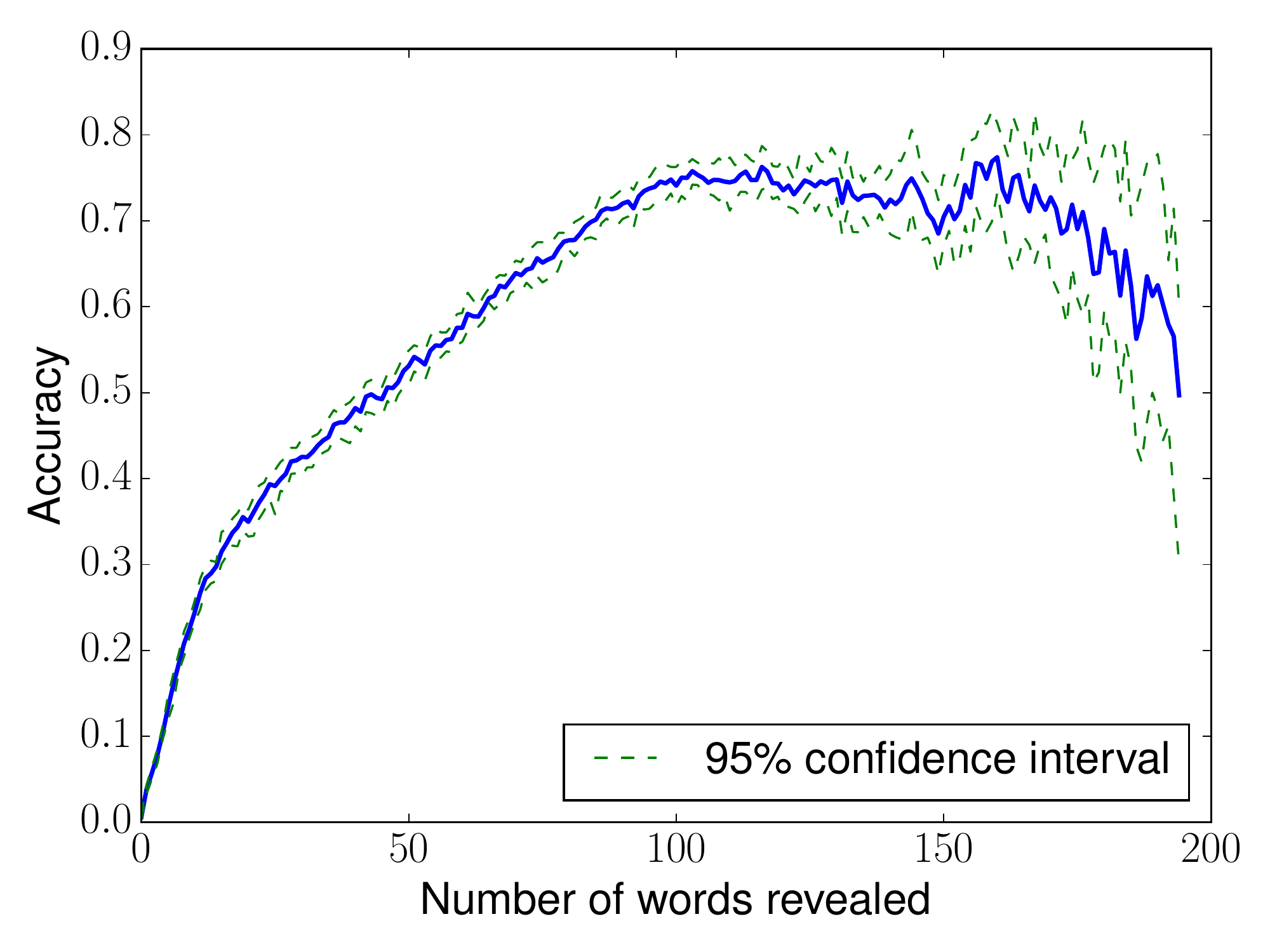}
\label{fig:content_model_acc}
}
\caption{Accuracy vs. the number of words revealed. (a) Real-time user
  performance. Each dot represents one user; dot size and color correspond to
  the number of questions the user answered. (b) Content model
  performance. Accuracy is measured based on predictions at each word.
  Accuracy improves as more words are revealed.}
\end{figure}

\paragraph{Implementation}

We collect question/answer pairs and log user buzzes from Protobowl, an online
multi-player quizbowl application.\footnote{\url{http://protobowl.com}}
Additionally, we include data from
\citet{Boyd-Graber:Satinoff:He:Daume-III-2012}.
Most buzzes are from strong tournament players.
After removing answers with
fewer than five questions and users who played fewer than twenty questions, we end up with
1045 answers, 37.7k questions and 3610 users.  We divide all questions into two
non-overlapping sets: one for training the content model and one for training
the buzzing policy.
The two sets are further divided into train/dev and train/dev/test sets randomly.
There are clearly two clusters of players
(Figure~\ref{fig:player_acc}): aggressive players who buzz early with varying
accuracies and cautious players who buzz late but maintain higher accuracy.
Our \gru{} content model (Figure~\ref{fig:content_model_acc}) is more accurate
with more input words---a behavior similar to human players.

Our input state must represent information from the content model and the
opponents.  Information from the content model takes the form of a \emph{belief
  vector}: a vector ($1\times 1045$) with the current estimate (as a log probability) of
each possible guess being the correct answer given our
input so far. We concatinate the belief vector from the previous time
step to capture sudden shifts in certainty, which are
often good opportunities to buzz.
In addition, we include the number of words seen and whether a wrong buzz has happened.

The opponent features include the number of questions the opponent has answered,
the average buzz position, and the error rate.  The basic \dqn{} has two hidden
layers, both with 128 hidden units.  The hidden layer for the opponent has ten
hidden units.  Similar to soccer, we experiment with two settings for
multitasking: (a) predicting how opponent buzzes; (b)
predicting the opponent type.  We approximate the ground truth for
(a) by $\min(1, t/\mbox{buzz position})$ and use the mean square error as the
loss function.  The ground truth for (b) is based on dividing players into four
groups according to their buzz positions---the percentage of question
revealed.

\begin{table*}[t]
\setlength{\tabcolsep}{1.5pt}

\begin{tabular}{lccc@{\hskip 3ex}ccc@{\hskip 3ex}ccc@{\hskip 3ex}ccc@{\hskip 3ex}ccc}
\toprule
\multirow{3}{*}{Model} & \multirow{2}{*}{Basic} & \multicolumn{2}{c}{Multitask} & \multicolumn{12}{c}{Basic vs. opponents buzzing at different positions (\%revealed (\#episodes))} \\
& & +action & +type & \multicolumn{3}{c}{\hspace{-3ex}$0-25\%$ (4.8k)} & \multicolumn{3}{c}{\hspace{-2ex}$25-50\%$ (18k)} & \multicolumn{3}{c}{\hspace{-2ex}$50-75\%$ (0.7k)} & \multicolumn{3}{c}{$75-100\%$ (1.3k)}  \\
\cmidrule(lr){2-4}\cmidrule(lr){5-16}
& \multicolumn{3}{c}{$R\uparrow$} & $R\uparrow$ & rush$\downarrow$ & miss$\downarrow$ & $R\uparrow$ & rush$\downarrow$ & miss$\downarrow$ & $R\uparrow$ & rush$\downarrow$ & miss$\downarrow$ & $R\uparrow$ & rush$\downarrow$ & miss$\downarrow$ \\
\midrule
\dron{}-concat & 1.04 & {\bf 1.34}$^\ast$ & {\bf 1.25} & -0.86 & 0.06 & 0.15 & 1.65 & 0.10 & 0.11 & -1.35 & 0.13 & 0.18 & 0.81 & 0.19 & 0.12 \\
\dronmoe{} & {\bf 1.29}$^\ast$ & 1.00 & {\bf 1.29}$^\ast$ & -0.46 & 0.06 & 0.15 & 1.92 & 0.10 & 0.11 & -1.44 & 0.18 & 0.16 & 0.56 & 0.22 & 0.10 \\
\dqn{}-world & 0.95 & - & - & -0.72 & 0.04 & 0.16 & 1.67 & 0.09 & 0.12 & -2.33 & 0.23 & 0.15 & -1.01 & 0.30 & 0.09 \\
\dqn{}-self & 0.80 & - & - & -0.46 & 0.09 & 0.12 & 1.48 & 0.14 & 0.10 & -2.76 & 0.30 & 0.12 & -1.97 & 0.38 & 0.07 \\
\bottomrule
\end{tabular}

\caption{Comparison between \dron{} and \dqn{} models. The left column shows the average reward of each model on the test set. The right column shows performance of the basic models against different types of players, including the average reward ($R$), the rate of buzzing incorrectly (rush) and the rate of missing the chance to buzz correctly (miss). $\uparrow$ means higher is better and $\downarrow$ means lower is better.
  In the left column, we indicate statistically significant results ($p<0.05$ in two-tailed pairwise $t$-tests) with boldface for vertical comparison and $^\ast$ for horizontal comparison.}
\label{tab:model_result}
\end{table*}

\paragraph{Results}

In addition to \dqn{}-world, we also compare with \dqn{}-self, a
baseline without interaction with opponents at all. \dqn{}-self is
ignorant of the opponents and plays the safe strategy:
answer as soon as the content model is confident.  During training, when the answer
prediction is correct, it receives reward 10 for \act{buzz} and -10 for
\act{wait}.  When the answer prediction is incorrect, it receives
reward -15 for \act{buzz} and 15 for \act{wait}.  Since all rewards
are immediate, we set $\gamma$ to 0 for \dqn{}-self.\footnote{This is equivalent to cost-sensitive classification.}
With data of the
opponents' responses, \dron{} and \dqn{}-world use the game payoff
(from the perspective of the computer) as the reward.

\begin{figure}
\centering
\includegraphics[width=0.6\linewidth]{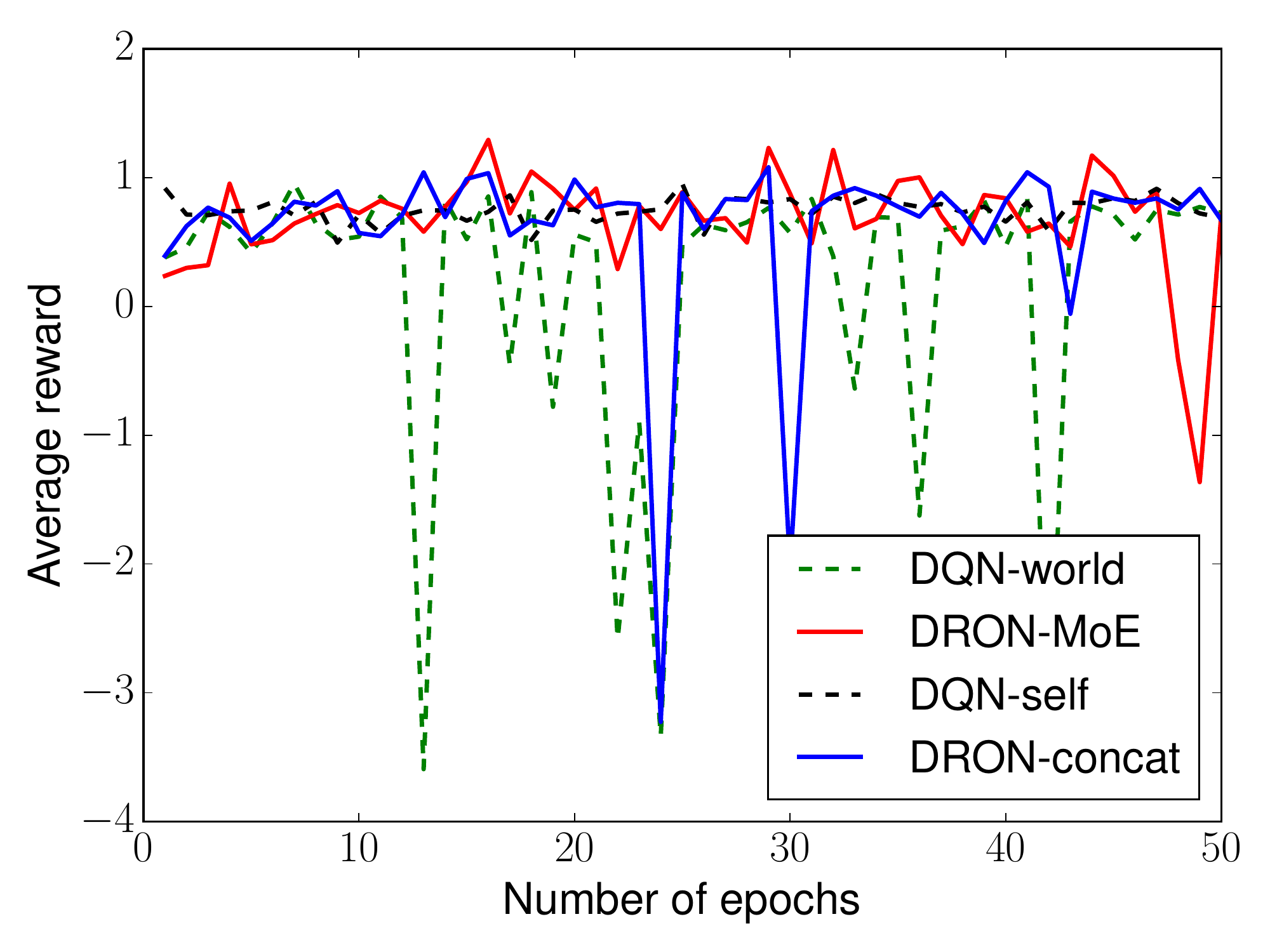}
\caption{Learning curves on Quizbowl over fifty epochs. \dron{} models are more
  stable than \dqn{}.}
\label{fig:qb_all_test_rewards}
\end{figure}

First we compare the average rewards on test set of our models---\dron{}-concat
and \dronmoe{} (with 3 experts)---and the baseline models: \dqn{}-self and
\dqn{}-world.  From the first column in Table~\ref{tab:model_result}, our models
achieve statistically significant improvements over the \dqn{} baselines and
\dronmoe{} outperforms \dron{}-concat.
In addition, the \dron{} models have much less variance compared to \dqn{}-world as the learning curves show in Figure~\ref{fig:qb_all_test_rewards}.

To investigate strategies learned by
these models, we show their performance against different types of players (as
defined at the end of ``Implementation'') in Table~\ref{tab:model_result}, right
column.  We compare three measures of
performance, the average reward ($R$), percentage of early and incorrect buzzes
(rush), and percentage of missing the chance to buzz correctly before the
opponent (miss).

All models beat Type~2 players, mainly
because they are the majority in our dataset.  As expected, \dqn{}-self learns a
safe strategy that tends to buzz early.  It performs the best against Type~1
players who answer early.  However, it has very high rush rate against cautious
players, resulting in much lower rewards against Type~3 and Type~4 players.  Without
opponent modeling, \dqn{}-world is biased towards the majority player, thus having
the same problem as \dqn{}-self when playing against players who buzz late.  Both
\dron{} models exploit cautious players while
holding their own against aggressive players.
Furthermore, \dronmoe{} matches \dqn{}-self against Type~1 players,
thus it discovers different buzzing strategies.

Figure~\ref{fig:buzz_example} shows an example question with buzz positions labeled.
The \dron{} agents demonstrate dynamic behavior against different players;
\dronmoe{} almost always buzzes right before the opponent in this example.
In addition, when the player buzzes wrong and the game continues, \dronmoe{} learns to wait longer since the opponent is gone, while the other agents are still in a rush.

As with the Soccer task,
adding extra supervision does not yield better results over \dronmoe{}
(Table~\ref{tab:model_result}) but significantly improves
\dron{}-concat.

\begin{figure}
\centering
\includegraphics[width=0.6\linewidth]{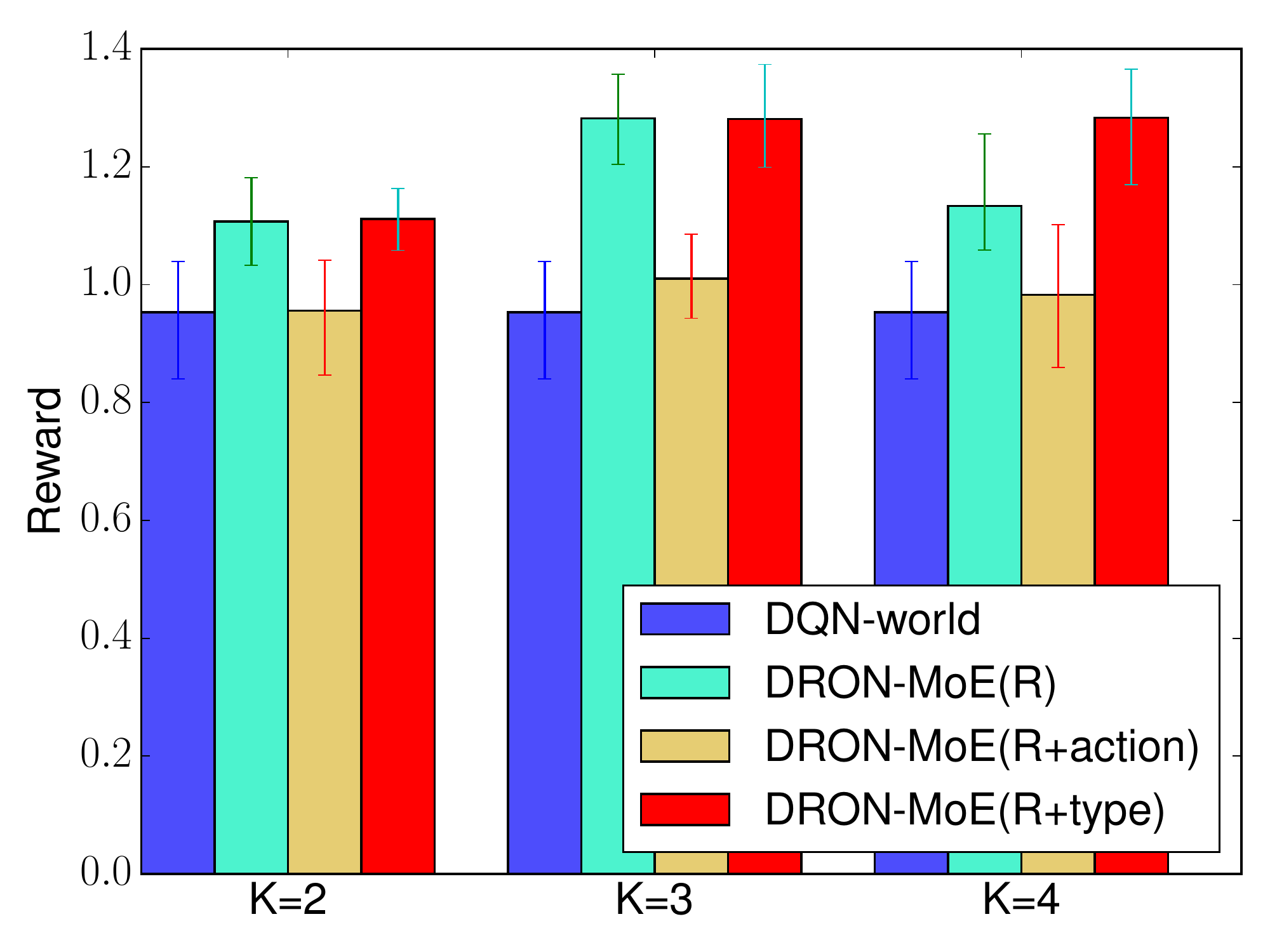}
\caption{Effect of varying the number experts (2--4) and multitasking on quiz bowl.  The
  error bars show the 90\% confidence interval.  \dronmoe{} consistently
  improves over \dqn{} regardless of the number of mixture components.
  Supervision of the opponent type is more helpful than the specific
  action taken.  }

\label{fig:qb_moe_bar}
\end{figure}

Figure~\ref{fig:qb_moe_bar} varies the number of experts in \dronmoe{}
($K$) from two to four.  Using a mixture model for the opponents
consistently improves over the \dqn{} baseline, and using three
experts gives better performance on this task.  For multitasking,
adding the action supervision does not help at all.  However, the more
high-level type supervision yields competent results, especially with
four experts, mostly because the number of experts matches the
number of types.

\section{Related Work and Discussion}
\label{sec:related_work}
\paragraph{Implicit vs. Explicit opponent modeling}

Opponent modeling has been studied extensively in games.  Most existing approaches fall into the category of explicit modeling, where a model (e.g., decision trees, neural networks, Bayesian models) is built to directly predict parameters of the opponent, e.g., actions~\cite{opponent-qlearning,game-theory-opponent-modeling}, private information~\cite{poker-opp,scrabble-opp}, or domain-specific strategies~\cite{schadd07opponentmodeling,bayesbluff}.  One difficulty is that the model may need a prohibitive number of examples before producing anything useful.  Another is that as the opponent behavior is modeled separately from the world, it is not always clear how to incorporate these predictions robustly into policy learning.  The results on multitasking \dron{} also suggest that improvement from explicit modeling is limited.  However, it is better suited to games of incomplete information, where it is clear what information needs to be predicted to achieve higher reward.

Our work is closely related to implicit opponent modeling.  Since the
agent aims to maximize its own expected reward without having to
identify the opponent's strategy, this approach does not have the
difficulty of incorporating
prediction of the opponent's parameters.  \citet{rubin11expert-imitator} and \citet{bard13implicit-modeling} construct a
a portfolio of strategies offline based on
domain knowledge or past experience for heads-up limit Texas hold'em; they then select strategies
online using multi-arm bandit algorithms.
Our approach does not have a clear online/offline distinction.
We learn strategies and their selector in a joint, probabilistic way.
However, the offline construction can be mimicked in our models by initializing expert networks with
\dqn{} pre-trained against different opponents.

\paragraph{Neural network opponent models}
\citet{davidson99opponent} applies neural networks to opponent modeling, where a simple multi-layer perceptron is trained as a classifier to predict opponent actions given game logs.
\citet{Lockett07evolvingexplicit} propose an architecture similar to \dron{}-concat that aims to identify the type of an opponent.
However, instead of learning a hidden representation, they learn a mixture weights over a pre-specified set of cardinal opponents;
and they use the neural network as a standalone solver without the reinforcement learning setting, which may not be suitable for more complex problems.
\citet{foerster16riddle} use modern neural networks to learn a group of parameter-sharing agents that solve a coordination task,
where each agent is controlled by a deep recurrent Q-Network~\cite{drqn}.
Our setting is different in that we control only one agent and the policy space of other agents is unknown.
Opponent modeling with neural networks remains understudied with ample room for improvement.

\section{Conclusion and Future Work}
\label{sec:conclusion}
Our general opponent modeling approach in the reinforcement learning setting incorporates (implicit) prediction of opponents' behavior into policy learning without domain knowledge.
We use recent deep Q-learning advances
to learn a representation of opponents that better maximizes available rewards.
The proposed network architectures are novel models that capture the interaction between opponent behavior and Q-values.
Our model is also flexible enough to include supervision for parameters of the opponents, much as in explicit modeling.

These gains can further benefit from advances in deep learning.  For example, \citet{eigen13dmoe} extends the Mixture-of-Experts network to a stacked model---deep Mixture-of-Experts---which can be combined with hierarchical reinforcement learning to learn a hierarchy of opponent strategies in large, complex domains such as online strategy games.
In addition, instead of hand-crafting opponent features, we can feed in raw opponent actions and use a recurrent neural network to learn the opponent representation.
Another important direction is to design online algorithms that can adapt to fast changing behavior and balance exploitation and exploration of opponents.

\section*{Acknowledgements}
We thank Hua He, Xiujun Li, and Mohit Iyyer for helpful discussions about deep Q-learning and our model. 
We also thank the anonymous reviewers for their insightful comments.
This work was supported by \abr{nsf} grant \abr{iis}-1320538.
Boyd-Graber is also partially supported by \abr{nsf} grants
\abr{ccf}-1409287 and \abr{ncse}-1422492. Any opinions, findings,
conclusions, or recommendations expressed here are those of the
authors and do not necessarily reflect the view of the sponsor.


\end{document}